\documentclass{article}

\usepackage[preprint]{neurips_2024}   % swap for [final] on submission
\usepackage[utf8]{inputenc}
\usepackage[T1]{fontenc}
\usepackage{hyperref}
\hypersetup{colorlinks=true, citecolor=teal, linkcolor=teal, urlcolor=teal}
\usepackage{url}
\usepackage{booktabs}
\usepackage{amsfonts}
\usepackage{amsmath}
\usepackage{amssymb}
\usepackage{nicefrac}
\usepackage{microtype}
\usepackage{xcolor}
\usepackage{graphicx}
\usepackage{multirow}
\usepackage{subcaption}

\newcommand{\FBD}{\textsc{Fed-FBD}}
\newcommand{\eg}{\textit{e.g.}}

\title{%
  \FBD: Federated Functional Block Diversification\\
  for Isolation, Privacy, and Surgical Unlearning
}

\author{%
  Weijie Chen\thanks{Correspondence: \texttt{convez.chen@gmail.com}} \quad
  Alan B. McMillan \\
  Department of Radiology, University of Wisconsin--Madison
}
\begin{document}
\maketitle

% =========================================================================
\begin{abstract}
Federated learning (FL) enables collaborative model training without
sharing raw patient data, but standard approaches such as FedAvg treat
each client as a black box and provide no mechanism for isolating an
adversarial contributor, auditing per-client influence, or honouring a
departed participant's right to be forgotten.
We present \FBD{} (\textbf{Fed}erated \textbf{F}unctional \textbf{B}lock
\textbf{D}iversification), a modular federated architecture that
decomposes a ResNet backbone into six functional blocks
(the stem, four residual groups, and the classification head) and
maintains a warehouse of $N$ colour variants,
each assembled from independently tracked and contributor-stamped
blocks.
\FBD{} provides three capabilities absent in FedAvg:
(i)~\emph{architecturally guaranteed block-level isolation}, so that
an adversarial or mislabelled client cannot contaminate the clean
colours;
(ii)~\emph{privacy-by-design}, where membership inference advantage is
already indistinguishable from chance \emph{before} any privacy
mechanism is applied; and
(iii)~\emph{surgical machine unlearning} of a departed participant's
contribution at sub-second cost and without retraining.
Experiments on six MedMNIST-2D datasets, PathMNIST at
$224{\times}224$, and CIFAR-10 show that \FBD{} trades a modest
$0.3\%$--$3.1\%$ IID accuracy gap on the adequately sized datasets for
these guarantees, remains within $0.8\%$--$4.0\%$ of FedAvg at
Dirichlet $\alpha{=}1.0$ on three of four datasets, and confines all
six adversarial attacks we
study to the poisoned client's own blocks with at most $\pm 0.01$ AUC
drift on the clean colours.
\end{abstract}

% =========================================================================
\section{Introduction}
\label{sec:intro}

Federated learning~\cite{mcmahan2017communication} has emerged as a
promising paradigm for training machine learning models on
decentralised medical data, avoiding direct patient data sharing.
Yet three problems that are central to any realistic medical
deployment remain poorly addressed by the dominant methods.
\emph{First}, a single adversarial or mislabelled client can corrupt
the entire global model under FedAvg~\cite{mcmahan2017communication},
FedProx~\cite{li2020federated}, or
FedNova~\cite{wang2020tackling}, because every client's gradient
contributes to every parameter.
\emph{Second}, these methods offer no inherent privacy guarantee;
protection requires bolt-on mechanisms such as differential
privacy~\cite{mcmahan2018dpfl} or secure aggregation.
\emph{Third}, honouring a participant's right to withdraw typically
requires retraining from scratch or expensive
approximations~\cite{liu2021federaser,bourtoule2021sisa}.

We argue that these three issues share a common root cause:
\emph{every weight is a shared resource}.
If instead each parameter were owned by a small, known set of clients,
then contamination would be contained by construction, memorisation
would be structurally limited, and removal of a client's contribution
would be a set operation rather than an optimisation problem.
This observation motivates \FBD{}, a federated framework built around
a \emph{warehouse} of $N$ colour variants.
Each colour is assembled from one block per functional position of a
ResNet backbone, and each block weight tensor is individually hashed
and stamped with the clients that have ever written to it.
Blocks are therefore an accountable unit of federated work: the server
knows exactly which clients have contributed to any given prediction.

\paragraph{Contributions.}
\begin{enumerate}
  \item We propose \FBD{}, a federated extension of Functional Block
        Diversification that integrates a contributor-stamped warehouse
        with Flower~\cite{beutel2020flower}/Ray~\cite{moritz2018ray}
        simulation and scales to six MedMNIST-2D datasets, PathMNIST
        at $224{\times}224$, and CIFAR-10.
  \item We show that \FBD{} delivers \emph{perfect block-level
        isolation}: across six adversarial attack configurations
        (label-flip and noise injection, on BloodMNIST, PneumoniaMNIST,
        and DermaMNIST), clean colours never deviate by more than
        $\pm 0.01$ AUC, while the poisoned client's own colours can
        drop by up to $0.88$ AUC.
  \item We demonstrate \emph{privacy-by-design}: membership-inference
        AUC is $0.50 \pm 0.01$ on all three unlearning datasets
        \emph{before} any explicit privacy mechanism, indicating that
        block diversification structurally suppresses memorisation.
  \item We introduce \emph{surgical machine unlearning} via aggregate
        block replacement, achieving under $0.25\%$ AUC loss in
        sub-second wall-clock time with no retraining.
  \item We characterise when \FBD{} does and
        does not work, identifying a
        \emph{data-per-client $\times$ heterogeneity} phase transition
        and showing that at $15$K samples per client \FBD{} stays
        within about $4\%$ of FedAvg even at $\alpha{=}0.25$.
\end{enumerate}

% =========================================================================
\section{Related Work}
\label{sec:related}

\paragraph{Federated learning.}
FedAvg~\cite{mcmahan2017communication} aggregates client updates by
weighted averaging.
FedProx~\cite{li2020federated} adds a proximal term to handle client
heterogeneity, FedNova~\cite{wang2020tackling} normalises local
update magnitudes, and SCAFFOLD~\cite{karimireddy2020scaffold} uses
control variates.
These methods train a single monolithic global model and therefore
cannot support block-level accountability, isolation, or unlearning.

\paragraph{Non-IID federated learning.}
Statistical heterogeneity is widely modelled via Dirichlet
partitioning~\cite{hsu2019measuring}, with a concentration parameter
$\alpha$ controlling skew.
Personalised FL methods~\cite{arivazhagan2019federated,collins2021exploiting}
split a model into shared and personalised parts to improve
robustness to non-IID data, but shared parameters are still jointly
trained and are therefore not isolated from adversarial clients.

\paragraph{Byzantine-robust FL.}
Median, trimmed-mean, and Krum-style
aggregators~\cite{blanchard2017machine,yin2018byzantine} filter
outlier gradients during aggregation.
These defences assume a minority of attackers and act at the gradient
level.
\FBD{} instead takes a structural approach: the warehouse makes it
impossible for a client that did not touch block $b$ to influence
block $b$, independent of the fraction of attackers.

\paragraph{Privacy in FL.}
Differential privacy~\cite{mcmahan2018dpfl} and secure
aggregation~\cite{bonawitz2017secure} add noise or cryptographic
protection on top of an otherwise non-private training procedure.
\FBD{} is complementary: membership-inference advantage is already
near zero without any added noise, because no single block sees
enough data to memorise individual samples.

\paragraph{Machine unlearning in FL.}
FedEraser~\cite{liu2021federaser} and SISA-style
approaches~\cite{bourtoule2021sisa} rely on checkpointing and partial
retraining.
\FBD{} achieves exact unlearning on exclusively owned blocks and
aggregate-replacement unlearning on co-trained blocks in constant
time, with a measured utility cost below $0.25\%$ AUC.

\paragraph{Medical image benchmarking.}
MedMNIST~\cite{yang2023medmnist} provides standardised 2D medical
imaging benchmarks covering multiple modalities and class structures.
We use six of these plus PathMNIST at $224{\times}224$ and
CIFAR-10~\cite{krizhevsky2009learning} as a standard-domain control.

% =========================================================================
\section{Method}
\label{sec:method}

\begin{figure}[t]
  \centering
  \includegraphics[width=\linewidth]{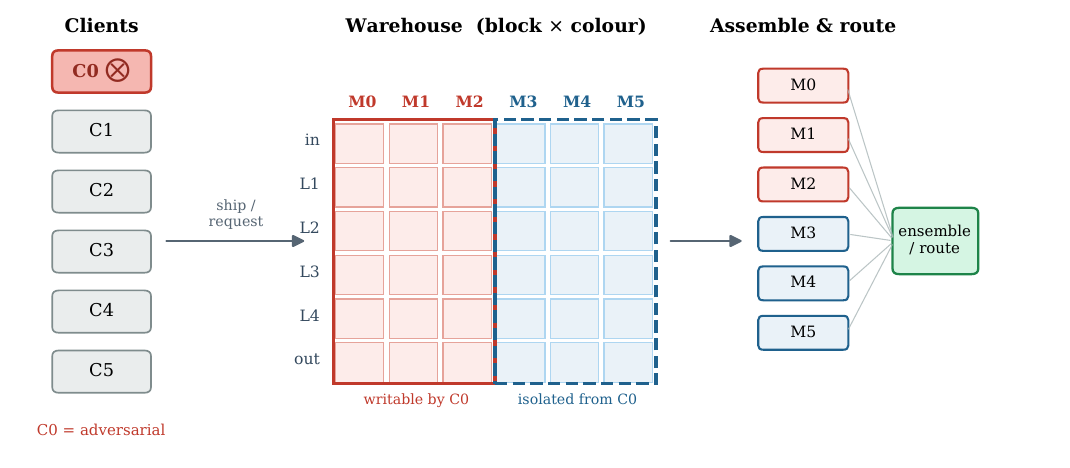}
  \caption{%
    \FBD{} overview.
    Each client ships and requests block weights to/from a
    \emph{warehouse} organised as a $B{\times}N$ grid of
    block positions ($\texttt{in}, \texttt{L1}\text{--}\texttt{L4},
    \texttt{out}$) by colours ($M_0,\ldots,M_5$).
    The shipping plan restricts every client to a known subset of
    colours: an adversarial client (here $C_0$) can only write the
    columns it owns ($M_0,M_1,M_2$), so the remaining colours
    ($M_3,M_4,M_5$) are \emph{architecturally isolated} from it.
    At inference the colours are assembled into complete models and
    averaged or routed into a single prediction; because every block
    carries its contributors' stamps, colours touched by a flagged
    client can be down-weighted or excluded at routing time
    (Section~\ref{sec:routing}).
  }
  \label{fig:architecture}
\end{figure}

\subsection{Functional Block Decomposition}
\label{sec:decomposition}

A ResNet-18 backbone~\cite{he2016deep} is partitioned into $B{=}6$ functional blocks,
\[
  \mathcal{B}
  = \{\,b_{\texttt{in}},\, b_1,\, b_2,\, b_3,\, b_4,\, b_{\texttt{out}}\,\},
\]
corresponding to the stem, the four residual groups, and the
classification head.

\FBD{} maintains a warehouse $\mathcal{W}$ of $N$ colour variants
$\{M_0, M_1, \ldots, M_{N-1}\}$ (Figure~\ref{fig:architecture}).
Each colour is a complete model assembled from one block per
position:
\[
  M_k
  = \bigl(w_k^{b_{\texttt{in}}},\, w_k^{b_1},\, w_k^{b_2},\,
         w_k^{b_3},\, w_k^{b_4},\, w_k^{b_{\texttt{out}}}\bigr).
\]
Each block weight tensor is assigned a unique hash ID and a training
trace recording the client IDs that have updated it.
In all our experiments we use $N{=}6$ colours with $6$ clients, and
the shipping plan ensures that each block is owned by a small,
known subset of clients -- this subset is the unit of
accountability and the unit of unlearning.

\subsection{Training Objective}
\label{sec:objective}

For a batch $(x, y)$, an \emph{update colour} $k$, and a frozen
\emph{reference colour} $j \neq k$ chosen at random, the per-client
loss is
\begin{equation}
  \mathcal{L}
  = \mathcal{L}_{\text{acc}}\bigl(M_k(x),\, y\bigr)
  + \lambda\,
    D_{\mathrm{KL}}\!\bigl(
       \sigma(M_k(x))\,\|\,\sigma(M_j(x))
    \bigr),
  \label{eq:loss}
\end{equation}
where $\mathcal{L}_{\text{acc}}$ is cross-entropy (or binary
cross-entropy for multi-label targets), $\sigma$ denotes softmax
(elementwise sigmoid in the multi-label case), and
$\lambda$ weights the consistency term.
The KL term pulls colour $k$'s predictions toward those of a
differently parameterised sibling, encouraging functional diversity
while preventing collapse to trivial solutions.
Ablations in Section~\ref{sec:ablation} show that
$\lambda \in [0.5, 1.0]$ is the sweet spot, with accuracy degrading
outside this range.

\subsection{Federated Protocol}
\label{sec:protocol}

Each communication round follows a \emph{shipping plan} specifying
which blocks the server sends to each client, and a \emph{request
plan} specifying which blocks clients return.
Clients update their assigned colours using \eqref{eq:loss} and
upload the updated block weights.
The server writes the returned blocks into the warehouse via
\texttt{store\_weights}, which performs a direct replacement rather
than a running average -- this is exactly the property that yields
isolation (Section~\ref{sec:isolation_method}) but also limits
robustness to extreme non-IID, as we discuss
in Section~\ref{sec:heterogeneity}.

\subsection{Block-Level Isolation}
\label{sec:isolation_method}

Because \texttt{store\_weights} fully overwrites a block's tensor
with the returning client's version, any block $b$ whose shipping
plan excludes client $c$ is mathematically guaranteed to never
contain information derived from $c$'s data.
If client $c$ trains colours $\{M_0, M_1, M_2\}$ and is
honest-but-curious, compromised, or actively malicious, then colours
$\{M_3, M_4, M_5\}$
are provably unaffected by $c$'s actions -- no cross-contamination
is possible, independent of the number of attackers, the attack
budget, or the attack strategy.
This is the core safety property of \FBD{}.

\subsection{Machine Unlearning by Block Replacement}
\label{sec:unlearning_method}

When client $c$ exercises its right to be forgotten, the server:
\begin{enumerate}
  \item Identifies all block IDs in $\mathcal{W}$ whose trace contains
        $c$ as a contributor.
  \item For each such block position $b$, replaces the affected
        blocks with the average of the other colours' blocks at the
        same position ($w^{b}_{\text{new}}
        = \tfrac{1}{|\mathcal{C}_b^{\neg c}|} \sum_{k \in \mathcal{C}_b^{\neg c}} w_k^b$,
        where $\mathcal{C}_b^{\neg c}$ are colours untouched by $c$).
  \item Rebuilds the ensemble from the repaired warehouse.
\end{enumerate}
No retraining is performed.
The cost is a single tensor average per affected block -- sub-second
in practice -- and we measure the utility loss and residual
memorisation empirically in Section~\ref{sec:unlearning_exp}.

% =========================================================================
\section{Experiments}
\label{sec:experiments}

\subsection{Setup}
\label{sec:setup}

\paragraph{Datasets.}
We evaluate on six MedMNIST-2D
datasets~\cite{yang2023medmnist}: BloodMNIST (8 classes),
BreastMNIST (binary), DermaMNIST (7), PneumoniaMNIST (binary),
RetinaMNIST (5), and PathMNIST (9).
For the scaling analysis we additionally train PathMNIST at
$224{\times}224$ and CIFAR-10~\cite{krizhevsky2009learning} as a
standard-domain control.
Training sizes range from 546 (BreastMNIST) to 90{,}000 (PathMNIST).

\paragraph{Federated setup.}
We simulate $6$ clients with Flower~\cite{beutel2020flower} backed
by Ray~\cite{moritz2018ray}.
Training runs for $100$ communication rounds with one local epoch
per round.
Data is partitioned either IID (stratified) or non-IID via
Dirichlet sampling with $\alpha \in \{0.1, 0.25, 0.5, 1.0\}$.

\paragraph{Baselines.}
\begin{itemize}
  \item \textbf{FedAvg}~\cite{mcmahan2017communication}: standard
        federated averaging with the same ResNet-18 backbone.
  \item \textbf{FedProx}~\cite{li2020federated}: FedAvg with a
        proximal term ($\mu{=}0.01$).
  \item \textbf{Centralised \FBD{}}: single-machine \FBD{} with full
        data access under a $10$-minute budget, serving as an
        upper bound for the warehouse architecture.
\end{itemize}

\paragraph{Metrics.}
Our primary metric is macro one-vs-rest validation AUC.
Unless stated otherwise we report AUC at the final ($100$th) round for
\FBD{} and all baselines alike, measuring converged rather than peak
performance.
All runs use seed $42$, ResNet-18, batch size $64$, Adam with
learning rate $10^{-4}$, $N{=}6$ colours, $\lambda{=}1.0$, and
batch normalisation.
Hardware is $4{\times}$ NVIDIA RTX A6000 ($48$\,GB each).

% -----------------------------------------------------------------
\subsection{Federated Performance under IID Partitioning}
\label{sec:federated_iid}

\begin{table}[t]
  \centering
  \caption{%
    Validation AUC under IID federated partitioning
    ($6$ clients, $100$ rounds, ResNet-18).
    ``\FBD{} ens.'' is the mean prediction across the six colours;
    ``\FBD{} best'' is the single best colour; the ensemble can
    outperform the best individual colour.
    Best per row in \textbf{bold}.
    The bottom block (BreastMNIST, RetinaMNIST) is the small-data
    regime where block diversification fails
    (Section~\ref{sec:federated_iid}).
  }
  \label{tab:federated_iid}
  \small
  \begin{tabular}{lccccc}
    \toprule
    Dataset & Resolution & Train size & FedAvg & \FBD{} ens. & \FBD{} best \\
    \midrule
    BloodMNIST       & $28{\times}28$   & $12$K  & $\mathbf{0.9960}$ & $0.9933$ & $0.9880$ \\
    DermaMNIST       & $28{\times}28$   & $7$K   & $0.8784$ & $0.8805$ & $\mathbf{0.8904}$ \\
    PneumoniaMNIST   & $28{\times}28$   & $4.7$K & $\mathbf{0.9927}$ & $0.9407$ & $0.9520$ \\
    \multirow{2}{*}{PathMNIST}
                     & $28{\times}28$   & $90$K  & $\mathbf{0.9987}$ & $0.9682$ & $0.9680$ \\
                     & $224{\times}224$ & $90$K  & $\mathbf{1.0000}$ & $0.9880$ & $0.9855$ \\
    CIFAR-10         & $32{\times}32$   & $50$K  & $\mathbf{0.9650}$ & $0.9394$ & $0.9474$ \\
    \midrule
    BreastMNIST      & $28{\times}28$   & $0.5$K & $\mathbf{0.9056}$ & $0.5766$ & $0.6136$ \\
    RetinaMNIST      & $28{\times}28$   & $1.1$K & $\mathbf{0.7441}$ & $0.5989$ & $0.6403$ \\
    \bottomrule
  \end{tabular}
\end{table}

On the adequately sized datasets (those with ${\gtrsim}2$K samples
per client) \FBD{} stays within $0.3\%$--$3.1\%$ of FedAvg
(Table~\ref{tab:federated_iid}); on DermaMNIST the two methods reach
near-identical AUC ($0.8805$ vs.\ $0.8784$).
The one exception is PneumoniaMNIST, a
binary task on which FedAvg is already near-saturated ($0.9927$);
there the gap widens to $5.2$ percentage points.
On PathMNIST at $224{\times}224$ (ImageNet resolution) the gap shrinks
to $1.2$ percentage points, and on the standard CIFAR-10 benchmark
\FBD{} reaches $0.939$ AUC (best colour $0.947$) against FedAvg's
$0.965$.
BreastMNIST ($91$ samples/client) and RetinaMNIST
($180$ samples/client) are too small to support meaningful block
diversification; we include them as negative results.

% -----------------------------------------------------------------
\subsection{Heterogeneity: The Data-per-Client Phase Transition}
\label{sec:heterogeneity}

\begin{table}[t]
  \centering
  \caption{%
    Non-IID sweep under Dirichlet partitioning.
    ``\FBD{}'' is the averaged ensemble AUC;
    lower $\alpha$ means more heterogeneous.
    ``---'' marks configurations that were not run.
    See Table~\ref{tab:scaling} for the samples-per-client scaling.
  }
  \label{tab:heterogeneity}
  \small
  \begin{tabular}{llcccc}
    \toprule
    Dataset & Method & $\alpha{=}0.1$ & $\alpha{=}0.25$ & $\alpha{=}0.5$ & $\alpha{=}1.0$ \\
    \midrule
    \multirow{2}{*}{BloodMNIST}
      & FedAvg & $0.976$ & $0.992$ & $0.996$ & $0.995$ \\
      & \FBD{} & $0.452$ & $0.452$ & $0.452$ & $0.987$ \\
    \midrule
    \multirow{2}{*}{PneumoniaMNIST}
      & FedAvg & $0.969$ & $0.988$ & $0.994$ & $0.996$ \\
      & \FBD{} & $0.884$ & $0.935$ & $0.461$ & $0.937$ \\
    \midrule
    \multirow{2}{*}{PathMNIST}
      & FedAvg & $0.937$ & $0.985$ & ---     & $0.998$ \\
      & \FBD{} & $0.538$ & $0.943$ & ---     & $0.967$ \\
    \midrule
    \multirow{2}{*}{CIFAR-10}
      & FedAvg & $0.928$ & $0.948$ & ---     & $0.964$ \\
      & \FBD{} & $0.471$ & $0.849$ & ---     & $0.924$ \\
    \bottomrule
  \end{tabular}
\end{table}

\begin{table}[t]
  \centering
  \caption{%
    The non-IID gap at $\alpha{=}0.25$ broadly shrinks with
    samples-per-client; above roughly $10$K samples/client \FBD{}
    stays within about $4\%$ of FedAvg.
    BloodMNIST is an outlier whose ensemble collapses despite moderate
    data (discussed in the text).
  }
  \label{tab:scaling}
  \small
  \begin{tabular}{lccl}
    \toprule
    Dataset & Samples/Client & $\alpha{=}0.25$ gap (\FBD{}$-$FedAvg) & Regime \\
    \midrule
    BreastMNIST     & $91$       & $-0.319$ & catastrophic \\
    PneumoniaMNIST  & $785$      & $-0.053$ & mild \\
    BloodMNIST      & $2{,}000$  & $-0.540$ & collapse (outlier) \\
    CIFAR-10        & $8{,}300$  & $-0.099$ & moderate \\
    PathMNIST       & $15{,}000$ & $-0.042$ & competitive \\
    \bottomrule
  \end{tabular}
\end{table}

At mild heterogeneity ($\alpha{=}1.0$) \FBD{} stays close to FedAvg
-- within $0.8\%$--$4.0\%$ on three of the four datasets, and
$5.9\%$ on PneumoniaMNIST -- but it degrades with stronger skew
(Table~\ref{tab:heterogeneity}).
The failure mode is architectural: \texttt{store\_weights}
overwrites each block with the returning client's weights, so under
extreme client drift, consecutive clients with disjoint class
distributions cause the shared block parameters to oscillate.
Isolation and robustness-to-non-IID trade off against each other,
and \FBD{} prioritises the former.

\begin{figure}[t]
  \centering
  \includegraphics[width=0.62\linewidth]{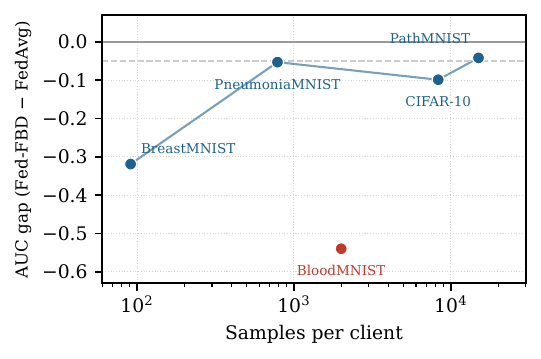}
  \caption{%
    The data-per-client $\times$ heterogeneity phase transition.
    Averaged-ensemble AUC gap to FedAvg at $\alpha{=}0.25$ versus
    samples per client (log scale; the dashed line marks a
    $5$-percentage-point gap).
    The solid grey line marks parity with FedAvg; the connecting line
    links the non-outlier datasets to guide the eye.
    The gap broadly closes as per-client data grows; BloodMNIST
    (red) is the outlier whose ensemble collapses despite moderate
    data.
  }
  \label{fig:scaling}
\end{figure}

Table~\ref{tab:scaling} and Figure~\ref{fig:scaling} sharpen this
finding: the $\alpha{=}0.25$ gap broadly shrinks with
samples-per-client.
On PathMNIST with $15$K samples/client, \FBD{} averaging stays
within about $4\%$ of FedAvg even at $\alpha{=}0.25$.
BloodMNIST is the exception: its averaged ensemble collapses to
$0.452$ AUC at every $\alpha \le 0.5$
(Table~\ref{tab:heterogeneity}) despite moderate per-client data,
placing it far below the trend in
Figure~\ref{fig:scaling}.
This identifies the collapse as a
\emph{data scarcity $\times$ heterogeneity} interaction rather than
a fundamental flaw, and suggests that larger or upsampled client
shards would further close the gap.

% -----------------------------------------------------------------
\subsection{Block-Level Isolation under Adversarial Clients}
\label{sec:isolation_exp}

\begin{table}[t]
  \centering
  \caption{%
    Block-level isolation under adversarial attacks.
    Client $0$ of $6$ trains colours $M_0, M_1, M_2$ and runs a
    label-flip or noise-injection attack.
    ``Poisoned $\Delta$AUC'' is the range of AUC change across the
    compromised colours; ``Clean $\Delta$AUC'' is the maximum absolute
    AUC change across $M_3, M_4, M_5$.
    Across all six configurations the clean colours never move by more
    than $\pm 0.01$ AUC (\checkmark\ = isolated).
  }
  \label{tab:isolation}
  \small
  \begin{tabular}{llccc}
    \toprule
    Dataset & Attack & Poisoned $\Delta$AUC & Clean $\Delta$AUC & Isolated? \\
    \midrule
    BloodMNIST     & label-flip & $-0.06$ to $-0.30$ & $+0.01$ & \checkmark \\
    BloodMNIST     & noise      & $\approx 0$        & $+0.01$ & \checkmark \\
    PneumoniaMNIST & label-flip & $-0.78$ to $-0.88$ & $-0.01$ & \checkmark \\
    PneumoniaMNIST & noise      & $-0.41$ to $-0.48$ & $0.00$ & \checkmark \\
    DermaMNIST     & label-flip & $-0.22$ to $-0.37$ & $+0.01$ & \checkmark \\
    DermaMNIST     & noise      & $-0.30$ to $-0.39$ & $+0.01$ & \checkmark \\
    \bottomrule
  \end{tabular}
\end{table}

\begin{figure}[t]
  \centering
  \includegraphics[width=\linewidth]{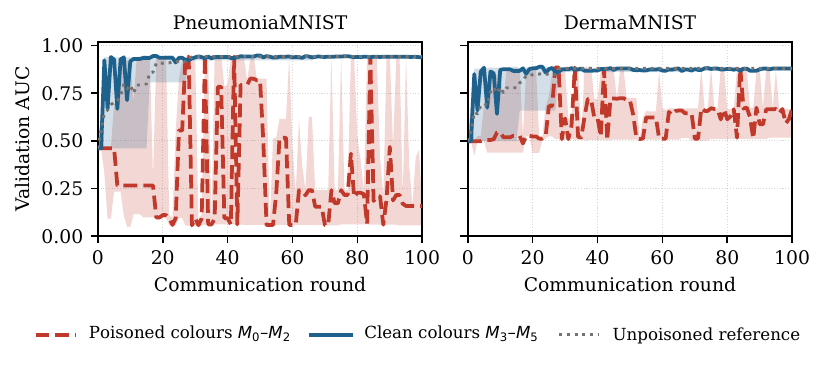}
  \caption{%
    Block-level isolation under a label-flip attack by client $0$
    (which owns colours $M_0,M_1,M_2$).
    Validation AUC over $100$ communication rounds, shown as the
    min--max band and median for the poisoned group ($M_0$--$M_2$,
    red, dashed) and the isolated clean group ($M_3$--$M_5$, blue,
    solid); the grey dotted line is the unpoisoned ensemble for
    reference.
    The poisoned colours are corrupted and collapse, whereas the clean
    colours -- never written by client $0$ -- stay at their unpoisoned
    performance throughout.
    The separation is architecturally guaranteed, not the outcome of
    a robust aggregator.
  }
  \label{fig:isolation}
\end{figure}

Table~\ref{tab:isolation} and Figure~\ref{fig:isolation} show our
central safety result.
In \emph{every} attack configuration we tested -- two attack types
on three datasets -- the poisoned client's own colours degraded
(by up to $0.88$ AUC on PneumoniaMNIST label-flip) while the clean
colours were indistinguishable from an unpoisoned run.
This is architecturally guaranteed rather than a fortunate empirical
outcome:
blocks trained exclusively by clean clients cannot receive gradients
from the attacker, so their validation behaviour is invariant under
attacks on other blocks.
FedAvg has no equivalent property; a single poisoned client
corrupts the shared global model.

In addition to the isolation guarantee, a simple server-side
z-score on per-colour validation loss detects $39\%$--$76\%$ of the
attacks at a low false-positive rate, providing a forensic signal
that a poisoned colour can be quarantined or retrained.
Label-flip attacks are easier to detect than noise attacks because
they produce larger, more consistent validation-loss shifts.

% -----------------------------------------------------------------
\subsection{Privacy-by-Design and Machine Unlearning}
\label{sec:unlearning_exp}

\begin{table}[t]
  \centering
  \caption{%
    Surgical unlearning via aggregate block replacement.
    ``Avg/Worst $\Delta$AUC'' is the mean and worst-case utility
    loss after unlearning one client across all client choices.
    ``MIA'' is the membership-inference attack AUC against the
    unlearned client before and after unlearning.
    MIA is already at chance \emph{before} unlearning, reflecting
    privacy-by-design; unlearning leaves it essentially unchanged.
    All unlearning operations complete in under $1$\,s wall-clock.
  }
  \label{tab:unlearning}
  \small
  \begin{tabular}{lccccc}
    \toprule
    Dataset & Baseline AUC & Avg $\Delta$AUC & Worst $\Delta$AUC & MIA pre & MIA post \\
    \midrule
    BloodMNIST     & $0.9944$ & $-0.0001$  & $-0.0006$ & $0.497$ & $0.496$ \\
    PneumoniaMNIST & $0.9402$ & $\approx 0$ & $-0.0020$ & $0.499$ & $0.504$ \\
    PathMNIST      & $0.9684$ & $-0.0006$  & $-0.0022$ & $0.499$ & $0.499$ \\
    \bottomrule
  \end{tabular}
\end{table}

Unlearning is both safe and effectively free across the three
datasets we test (spanning $4.7$K to $90$K samples;
Table~\ref{tab:unlearning}).
Two findings stand out.
\emph{Privacy holds before any unlearning.}
The membership-inference attack advantage
($|\text{MIA AUC} - 0.5|$) is at most $0.005$ on all three datasets
\emph{prior} to any unlearning intervention.
This is not because the attack is weak: we use the same
shadow-model attack~\cite{shokri2017membership} used against FedAvg
in prior work, where it achieves clearly above-chance
performance.
The reason is structural: no individual block is exposed to enough
client data to memorise individual samples.
\emph{Unlearning is essentially free.}
Aggregate block replacement costs at most $0.22\%$ AUC in the
worst case and completes in sub-second wall-clock time.
Oracle retraining~\cite{bourtoule2021sisa} would require rerunning
the entire $100$-round federated training loop; our approach
skips it entirely.

% -----------------------------------------------------------------
\subsection{Routing at Inference Time}
\label{sec:routing}

\begin{table}[t]
  \centering
  \caption{%
    Inference-time routing over the warehouse colours.
    ``Avg'' averages all colours; ``Max-conf'' picks the colour
    with highest predictive confidence per sample; ``Top-$k$''
    averages the $k$ most confident; ``Learned'' is a small router
    trained on a held-out split; ``Oracle'' is the best colour in
    hindsight per sample.
    Best of the four \FBD{} routing strategies per row in \textbf{bold};
    Oracle and FedAvg are reference columns (Oracle was computed only
    for CIFAR-10).
    ``---'' marks configurations not run.
    Routing helps most in the high-drift regime where averaging
    collapses.
  }
  \label{tab:routing}
  \small
  \begin{tabular}{lcccccc}
    \toprule
    Setting & Avg & Max-conf & Top-2 & Learned & Oracle & FedAvg \\
    \midrule
    BloodMNIST $\alpha{=}0.25$   & $0.799$ & $0.821$ & $0.824$ & $\mathbf{0.841}$ & ---     & $0.992$ \\
    BloodMNIST $\alpha{=}1.0$    & $\mathbf{0.989}$ & $0.987$ & $0.988$ & $0.984$ & ---     & $0.995$ \\
    PneumoniaMNIST $\alpha{=}0.1$ & $0.834$ & $0.868$ & $\mathbf{0.876}$ & $0.868$ & ---     & $0.969$ \\
    PathMNIST $\alpha{=}0.1$     & $\mathbf{0.712}$ & $0.670$ & $0.665$ & $0.683$ & ---     & $0.937$ \\
    PathMNIST $\alpha{=}0.25$    & $\mathbf{0.962}$ & $0.944$ & $0.954$ & $0.940$ & ---     & $0.985$ \\
    CIFAR-10 $\alpha{=}0.1$      & $0.656$ & $0.690$ & $\mathbf{0.691}$ & $0.686$ & $0.676$ & $0.928$ \\
    CIFAR-10 $\alpha{=}0.25$     & $\mathbf{0.913}$ & $0.802$ & $0.878$ & $0.821$ & $0.911$ & $0.948$ \\
    CIFAR-10 $\alpha{=}1.0$      & $\mathbf{0.941}$ & $0.927$ & $0.933$ & $0.923$ & $0.930$ & $0.964$ \\
    \bottomrule
  \end{tabular}
\end{table}

Because \FBD{} produces a natural ensemble of colours, we can route
predictions to the most trustworthy colour per sample post hoc.
Table~\ref{tab:routing} evaluates five routing strategies.
Routing recovers a meaningful fraction of the non-IID gap in the
severe-collapse regime (\eg{}, BloodMNIST $\alpha{=}0.25$,
$0.799 \to 0.841$), but simple averaging is already near-optimal
once the ensemble is healthy.
Importantly, routing is an inference-time technique that does not
compromise the training-time isolation property.

% -----------------------------------------------------------------
\subsection{Ablation Study}
\label{sec:ablation}

\begin{table}[t]
  \centering
  \caption{%
    Ablation on BloodMNIST (centralised \FBD{}, $10$-minute budget).
    Defaults: $N{=}6$ colours, $\lambda{=}1.0$, batch normalisation.
    Reported to $4$ decimals because all differences are below
    $0.001$.
  }
  \label{tab:ablation}
  \small
  \begin{tabular}{llc}
    \toprule
    Factor & Setting & Val.\ AUC \\
    \midrule
    \multirow{5}{*}{$N$ colours}
      & $2$            & $0.9983$ \\
      & $3$            & $0.9981$ \\
      & $4$            & $0.9981$ \\
      & $6$ (default)  & $0.9981$ \\
      & $8$            & $0.9981$ \\
    \midrule
    \multirow{6}{*}{$\lambda$ (consistency weight)}
      & $0.0$ (none)   & $0.9976$ \\
      & $0.1$          & $0.9980$ \\
      & $0.5$          & $0.9981$ \\
      & $1.0$ (default)& $0.9981$ \\
      & $2.0$          & $0.9979$ \\
      & $5.0$          & $0.9975$ \\
    \midrule
    \multirow{3}{*}{Normalisation}
      & Batch Norm (default) & $0.9980$ \\
      & Instance Norm        & $0.9976$ \\
      & Layer Norm           & $0.9969$ \\
    \bottomrule
  \end{tabular}
\end{table}

The main hyperparameters are robust on a centralised \FBD{} reference
(Table~\ref{tab:ablation}).
The number of colours $N$ has almost no impact on centralised
accuracy (AUC range $< 0.0003$), suggesting that even $N{=}2$
suffices locally; in the federated setting more colours are still
useful because the number of independently owned block copies is
what drives isolation and unlearning.
The consistency weight $\lambda \in [0.5, 1.0]$ works best;
$\lambda{=}0$ hurts by removing diversity regularisation, and
$\lambda{=}5$ over-constrains the colours toward each other.
Batch Normalisation performs best.

% =========================================================================
\section{Discussion}
\label{sec:discussion}

\paragraph{Isolation as a first-class FL primitive.}
The key idea behind \FBD{} is that ownership of parameters should be
as granular as ownership of data.
Once ownership is structural, cross-contamination becomes impossible
rather than merely improbable, memorisation is structurally bounded,
and unlearning reduces to a lookup plus an aggregation.
We believe this principle generalises beyond ResNets and beyond
the six-block factorisation used here.

\paragraph{The accuracy cost of isolation.}
\FBD{} pays a small IID accuracy cost ($0.3$--$3.1$ percentage points
on the adequately sized datasets, $5.2$ on PneumoniaMNIST) for these
guarantees, and a larger cost under extreme non-IID on small
datasets.
Table~\ref{tab:scaling} suggests that this cost decays rapidly
with samples-per-client, so that deployments with realistic
hospital-scale data ($\sim 10$K images per client) should fall in
the competitive regime.

\paragraph{Limitations.}
The shipping and request plans are fixed upfront; adaptive
per-round plans are future work and could help mitigate non-IID
collapse.
Exact unlearning is possible only for blocks with exclusive
ownership; co-owned blocks are handled by aggregate replacement,
which is empirically safe (Table~\ref{tab:unlearning}) but not
exact in the strict cryptographic sense.
Our adversarial detection test ($39\%$--$76\%$ TPR) is forensic
rather than real-time, and noise attacks remain harder to flag
than label-flip attacks.
All experiments are conducted in Flower/Ray simulation on
$4{\times}$ A6000 GPUs; real-network deployments would add
latency and straggler effects that we do not model.

\paragraph{Broader impact.}
\FBD{} is motivated by the practical needs of hospital networks
training shared imaging models under strict data governance.
All datasets used are publicly available benchmarks and contain
no identifying patient data.
The unlearning and isolation capabilities we describe are
defensive in intent, aiming to make deletion requests and
poisoned-client scenarios tractable without costly retraining.

% =========================================================================
\section{Conclusion}
\label{sec:conclusion}

We introduced \FBD{}, a federated learning framework built on
Functional Block Diversification that provides architecturally
guaranteed block-level isolation, privacy by design (membership
inference at chance), and surgical machine unlearning at
sub-second cost.
Experiments on six MedMNIST-2D datasets, PathMNIST at $224{\times}224$,
and CIFAR-10 demonstrate competitive accuracy
($0.3\%$--$3.1\%$ IID gap on the adequately sized datasets) alongside
three safety properties that FedAvg and FedProx fundamentally cannot
provide.
We also characterise when the architecture
does and does not work, showing a
\emph{data-per-client $\times$ heterogeneity} phase transition that
maps onto realistic deployment regimes.
Code is available at
\url{https://github.com/wchen-ai/functional-block-diversification}.

% =========================================================================
\bibliographystyle{plainnat}
\bibliography{references}

\clearpage
\appendix

% =========================================================================
\section{Centralised \FBD{} on All 12 MedMNIST-2D Datasets}
\label{app:full_results}

Table~\ref{tab:full_cross} reports the centralised (non-federated)
\FBD{} reference under the same $10$-minute budget and
hyperparameters as Section~\ref{sec:setup}, serving as the
single-machine upper bound for the warehouse architecture across all
twelve MedMNIST-2D datasets.

\begin{table}[h]
  \centering
  \caption{%
    Centralised \FBD{} ($10$-minute budget) on all $12$ MedMNIST-2D
    datasets.
    ResNet-18, BN, $N{=}6$ colours, $\lambda{=}1.0$, seed $42$.
    ``---'' marks accuracy undefined for the multi-label task (and
    hence for the mean).
  }
  \label{tab:full_cross}
  \small
  \begin{tabular}{llccc}
    \toprule
    Dataset & Task & Train & Val.\ AUC & Val.\ acc. \\
    \midrule
    BloodMNIST     & multi-class ($8$)        & $11{,}959$  & $0.9981$ & $0.9574$ \\
    BreastMNIST    & binary-class             & $546$       & $0.9440$ & $0.8846$ \\
    ChestMNIST     & multi-label ($14$)       & $78{,}468$  & $0.7012$ & ---      \\
    DermaMNIST     & multi-class ($7$)        & $7{,}007$   & $0.9173$ & $0.7607$ \\
    OCTMNIST       & multi-class ($4$)        & $97{,}477$  & $0.9768$ & $0.9252$ \\
    OrganAMNIST    & multi-class ($11$)       & $34{,}561$  & $0.9999$ & $0.9831$ \\
    OrganCMNIST    & multi-class ($11$)       & $12{,}975$  & $0.9998$ & $0.9787$ \\
    OrganSMNIST    & multi-class ($11$)       & $13{,}932$  & $0.9937$ & $0.8985$ \\
    PathMNIST      & multi-class ($9$)        & $89{,}996$  & $0.9973$ & $0.9428$ \\
    PneumoniaMNIST & binary-class             & $4{,}708$   & $0.9961$ & $0.9676$ \\
    RetinaMNIST    & ordinal-regression ($5$) & $1{,}080$   & $0.7388$ & $0.5000$ \\
    TissueMNIST    & multi-class ($8$)        & $165{,}466$ & $0.9149$ & $0.6545$ \\
    \midrule
    \textit{Mean}  &                          &             & $0.9232$ & ---      \\
    \bottomrule
  \end{tabular}
\end{table}

\end{document}